\documentclass{article}
\usepackage{spconf,amsmath,epsfig,amssymb}
\usepackage{xcolor}
\usepackage{booktabs}
\usepackage{array,multirow,graphicx}
\usepackage{multicol}
\usepackage{tikz}
\usepackage{caption}
\usepackage{subcaption}
\usetikzlibrary{tikzmark,decorations.pathreplacing,positioning}
\tikzstyle{curly} = [decorate,decoration={brace,amplitude=10pt}]

\newcommand{\nada}[1]{}

\definecolor{Cyan}{rgb}{0,.68,.94} %
\usepackage[abs]{overpic}  %
\newcommand{\overimg}[3][]{%
    \begin{overpic}[#1]{#2}%
      \put (0, 2) {%
        \setlength{\fboxsep}{2pt}%
        \colorbox{Cyan!0!white}{%
          \scriptsize\sffamily\vphantom{y}%
          #3%
        }%
      }%
    \end{overpic}%
}

\title{%
On The Role of Alias and Band-Shift for Sentinel-2 Super-Resolution
}

\name{~~~\hfill Ngoc Long Nguyen ${^{a,*}}$ \hfill J\'er\'emy Anger $^{a,b,*}$ \hfill Lara Raad $^{c,*}$ \hfill Bruno Galerne $^{d,e}$ \hfill Gabriele Facciolo $^{a}$ \hfill~~~}
\address{%
\normalsize $^a$ Université Paris-Saclay, CNRS, ENS Paris-Saclay, Centre Borelli, France \hspace{0.6cm}
\normalsize $^b$ Kayrros SAS \\ 
\normalsize $^c$ Univ Gustave Eiffel, CNRS, LIGM, F-77454 Marne-la-Vallée, France \\
\normalsize $^d$ Institut Denis Poisson,  Universit\'e d'Orl\'eans, Université de Tours, CNRS, France 
~~\normalsize $^e$ Institut universitaire de France (IUF) \\
\thanks{* These authors contributed equally to this work. This work was supported by a grant from Région Île-de-France, and DGA Astrid Maturation project SURECAVI ANR-21-ASM3-0002.
This work was performed using HPC resources from GENCI–IDRIS (grants 2022-AD011012453R1, 2022-AD011012458R1 and 2022-AD011012472R1).}%
}

\begin{document}
\maketitle
\begin{abstract}
In this work, we study the problem of single-image super-resolution (SISR) of Sentinel-2 imagery.
We show that thanks to its unique sensor specification, namely the inter-band shift and alias, that deep-learning methods are able to recover fine details. By training a model using a simple $L_1$ loss, results are free of hallucinated details.
For this study, we build a dataset of pairs of images Sentinel-2/PlanetScope to train and evaluate our super-resolution (SR) model.
\end{abstract}
\section{Introduction}
\label{sec:intro}
\begin{figure}[h!]
    \def\s{0.325\linewidth}
    \centering
    \foreach \a in {suburbs,parking,airport,tanks2,tanks}{
    \foreach \b in {s2,Grgb,ps}{
        \includegraphics[width=\s]{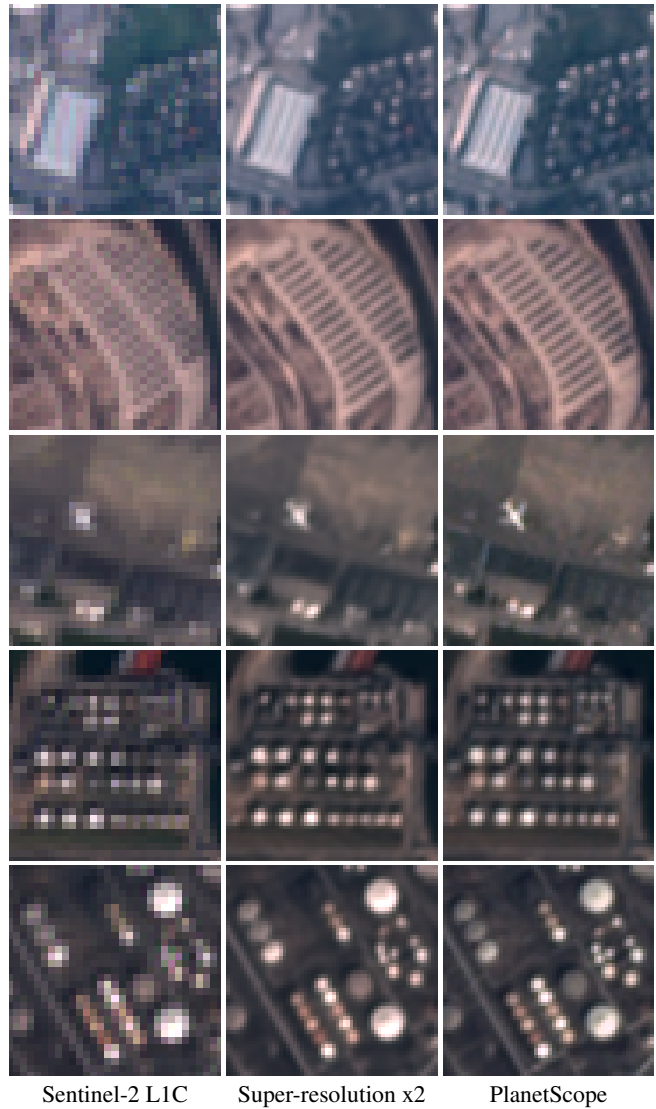}%
    }\\[0.1em]
    }
    \begin{subfigure}{\s}
        \caption*{Sentinel-2 L1C}
    \end{subfigure}
    \begin{subfigure}{\s}
        \caption*{Super-resolution x2}
    \end{subfigure}
    \begin{subfigure}{\s}
        \caption*{PlanetScope}
    \end{subfigure}
    \vspace{-0.8em}\caption{SISR results obtained with the $L_1$ loss. We argue that the characteristic alias and band-shift are key for x2 SR of Sentinel-2 imagery.}\label{fig:results}
\end{figure}

The use of satellite imagery has become increasingly prevalent in a variety of fields, from environmental monitoring to urban planning. One such satellite is the Sentinel-2 constellation, which provides recurrent 10m/pixel resolution optical imagery.
The high frequency revisit of Sentinel-2 makes it useful for monitoring temporal changes, such as the growth of crops or the spread of urban development. However, the relatively low spatial resolution can be a limitation for certain applications, such as identifying small objects or analyzing fine-scale features.

In this paper, we propose a deep learning approach for SISR of Sentinel-2 imagery. Unlike previous methods that aim for a x4 increase in spatial resolution, our work focuses on a x2 increase, which we argue is a more reasonable and practical choice. 
Additionally, we avoid using generative adversarial networks (GANs) in our method, as they have been known to introduce hallucinations and artifacts that can be undesirable for sensitive applications.
Instead, we use an $L_1$ cost function, which has been shown to effectively preserve image details while minimizing distortion~\cite{Blau_2018_CVPR_short} (see Figure~\ref{fig:results}).

This study focuses on understanding what makes SISR of Sentinel-2 imagery possible.
To this aim, we explore two unique characteristics of Sentinel-2: the alias and the inter-band shift and find that they enable the reconstruction of fine structures.
It is worth noting that super-resolving the 10m bands of Sentinel-2 is a relatively new problem, and while we do not aim to achieve the best possible results, we provide an analysis of the specific features of Sentinel-2 imagery that are relevant for SR.

\begin{figure}
    \centering
    \def\s{0.325\linewidth}
    \begin{subfigure}{\s}
        \includegraphics[width=\linewidth]{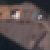}\caption{S2 (input)}
    \end{subfigure}
    \begin{subfigure}{\s}
        \overimg[width=\linewidth]{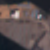}{\input{img/many-results/worksite-bicubic.tex}dB}\caption{Bicubic}
    \end{subfigure}
    \begin{subfigure}{\s}
        \overimg[width=\linewidth]{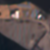}{\input{img/many-results/worksite-Gr_g_b.tex}dB}\caption{$L_1$ \small (per channel)}
    \end{subfigure}
    \\[0.1em]
    \begin{subfigure}{\s}
        \overimg[width=\linewidth]{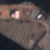}{\input{img/many-results/worksite-Grgb_gan_pretrained.tex}dB}\caption{GAN loss}
    \end{subfigure}
    \begin{subfigure}{\s}
        \overimg[width=\linewidth]{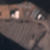}{\input{img/many-results/worksite-Grgb.tex}dB}\caption{$L_1$ loss}
    \end{subfigure}
    \begin{subfigure}{\s}
        \includegraphics[width=\linewidth]{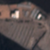}\caption{PS (ground-truth)}
    \end{subfigure}
    \caption{SR results from different models. The per-channel result corresponds to one model per channel trained independently. The GAN and the $L_1$ loss are able to restore similar details.}
    \label{fig:results-worksite}
\end{figure}

\section{Related work}
\label{sec:related-work}

Early research on SISR of Sentinel-2 images focuses on pan-sharpening the lower-resolution (20m and 60m) bands to complete a uniform 10m GSD data cube~\cite{lanaras2018super_short, gargiulo2019fast_short}. Recent trends are super-resolving the 10m bands of Sentinel-2 using other relevant very high-resolution satellites. For example, \cite{pineda2020generative_short} generates low-resolution/high-resolution (LR-HR) image pairs from the PeruSat-1 satellite (2.8m GSD) to train a x4 SR model and use it to reconstruct fine textures in the Sentinel-2 10m bands. However, these techniques use a pre-determined degradation model, like bicubic downsampling, to create LR from HR. So when the input deviates from the pre-defined degradation model, the performance may drop substantially. To fill the gap between simulated and real-world remote sensing images, real HR satellites such as PlanetScope~\cite{galar2020super_short, zabalza2022super_short}, VEN\textmu S~\cite{michel2022sen2venmus}, and WorldView~\cite{salgueiro2020super_short} are used directly to supervise the SR of Sentinel-2.
Perceptual losses, such as GAN or high-level feature matching, are used in these works to produce sharp outputs. 
More recently, \cite{Nguyen_2023_L1BSR} proposes a self-supervised super-resolution for Sentinel-2 L1B products, which are not available for historical data.

Besides focusing on perceptual restoration, most past studies do not justify why the reconstruction of actual high-frequency details is feasible from a single multi-band image.
Results reported in~\cite{nguyen2021self_short} suggest that alias and displacement between frames are crucial for exploiting complementary information in different frames (or different spectral bands in our case) and obtaining up-to-par SR performance.

\section{Method}
\label{sec:method}

Our method is specifically designed for x2 SR of Sentinel-2 images.
It is based on the ESRGAN architecture~\cite{Wang_et_al_ESRGAN_2018_short}, adapted for a smaller network and a single-term loss.
The model is trained on pairs of Sentinel-2 and PlanetScope images, suitable for x2 SR, as described in Section~\ref{sec:dataset}.

\medskip\noindent\textbf{Architecture.}
Given that ESRGAN was developed for a x4 SR factor, we propose some adjustments for a factor of 2.
We found that using only 8 RRDB blocks instead of 23 RRDB blocks~\cite{Wang_et_al_ESRGAN_2018_short} was enough to obtain satisfactory results while significantly reducing the training and inference time.

\medskip\noindent\textbf{Cost function.}
The ESRGAN model~\cite{Wang_et_al_ESRGAN_2018_short}, was initially trained on a set of HR natural images from the DIV2K dataset~\cite{Agustsson_2017_CVPR_Workshops_short}.
It uses a base model trained with a loss $L_1$, followed by a second training phase using a cost function that includes the relativistic discriminator loss~\cite{jolicoeur2018relativistic_short}, the perceptual loss~\cite{10.1007/978-3-319-46475-6_43_short}, and the $L_1$ loss.
However, when adapting the model for Sentinel-2 SR, we found that training the model on Sentinel-2/PlanetScope image pairs using only the $L_1$ loss instead of the complete loss function with perceptual terms resulted in a similar detail reconstruction (second row in Figure~\ref{fig:results-worksite}).
This suggests that, thanks to the alias and inter-band shift present in Sentinel-2 imagery, the problem is better posed. Hence the $L_1$ loss is sufficient for successful x2 SR.
This is further explored in Section~\ref{subsec:real-training}.

\begin{figure*}[t!]
\centering
\setlength{\tabcolsep}{1.3pt}
\begin{tabular}{cccccccc}
    \includegraphics[width=0.138\linewidth]{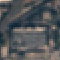}&
    \includegraphics[width=0.138\linewidth]{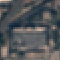}&
    \includegraphics[width=0.138\linewidth]{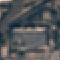}&
    \includegraphics[width=0.138\linewidth]{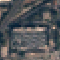}&
    \includegraphics[width=0.138\linewidth]{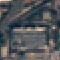}&
    \includegraphics[width=0.138\linewidth]{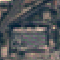}&
    \\
    \includegraphics[width=0.138\linewidth]{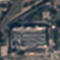}&
    \includegraphics[width=0.138\linewidth]{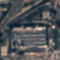}&
    \includegraphics[width=0.138\linewidth]{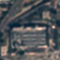}&
    \includegraphics[width=0.138\linewidth]{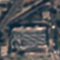}&
    \includegraphics[width=0.138\linewidth]{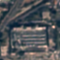}&
    \includegraphics[width=0.138\linewidth]{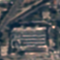}&
    \includegraphics[width=0.138\linewidth]{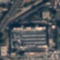}\\[0.2em]

    no shift & fixed shift & random shift & no shift & fixed shift & random shift & ground-truth &
    \\~\\
\end{tabular}
\begin{tikzpicture}[yshift=1.1\baselineskip,overlay,remember picture]
    \draw (-1.5,0) -- (-8.6,0) node[midway,below=0.0] {no alias};
    \draw (6.2,0) -- (-1.0,0) node[midway,below=0.0] {alias};
\end{tikzpicture}
    \vspace{-0.8em}\caption{Synthetic dataset. The top row shows simulated LR images corresponding to the six acquisition configurations, and the bottom row shows the SR results obtained with networks trained on these specific synthetic datasets. The bottom-right image is the ground-truth HR.}
\label{fig:synthetic-dataset}
\end{figure*}
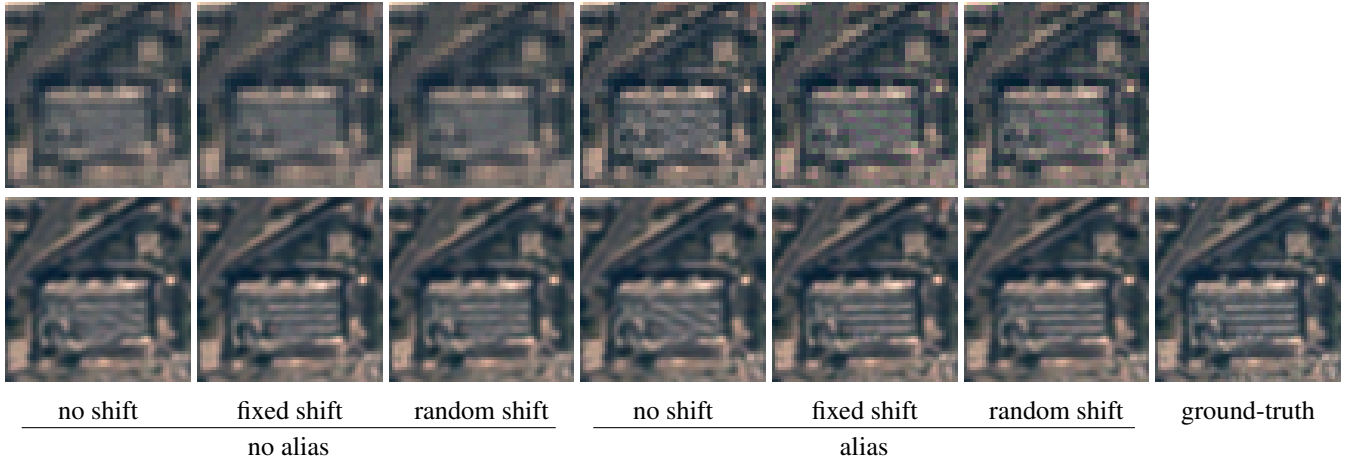

\section{S2/PS DATASET}
\label{sec:dataset}

For this study, we built a dataset of Sentinel-2 L1C and PlanetScope image pairs, referred to as the S2/PS dataset. The Sentinel-2 L1C images have a spatial resolution of 10m/px, while the PlanetScope images have a resolution of 3m/px. The latter were resampled to 5m/px and registered with the S2 coordinate system with bicubic interpolation.
Since PlanetScope images tend to be well-sampled~\cite{anger2019assessing_short}, the resampling step does not introduce a significant loss of information.
The pairs are from images taken on the same day, a few hours apart, to minimize changes due to vegetation or human activity. The remaining changes in the images include the presence of clouds, differences in satellite perspective, and shadows.
In this work, we use the PlanetScope images acquired with the PS2 instrument (Dove Classic), so we restrict our study to the three visible bands (blue, green, and red).

To prepare this dataset, we performed equalization of the mean and standard deviation for each band from each PlanetScope image to the corresponding Sentinel-2 image.
The residual spatial shift between the  downsampled PlanetScope image and the Sentinel-2 one is estimated using the phase correlation algorithm. Then, the PlanetScope image is resampled using a third-order spline interpolation and introducing zeros where information was missing.
Pairs with a phase correlation score below 0.55 were removed from the dataset.

We used 380 full-scene images and extracted up to 20 LR crops  of size $200\times200$ from each image.
The test set consists of 65 of these scenes (or 693 crops) and is geospatially disjoint from the train set (3680 crops). The validation set (406 crops) is selected from different dates.
Figure~\ref{fig:results} shows zoom-in crops from the dataset, where alias and inter-band shift are clearly visible in the S2 images.

\begin{table}[t!]
    \centering
    \begin{tabular}{llccc}
    \toprule
     & & \multicolumn{3}{c}{PSNR (dB)}\tabularnewline
     & & Test set & Train set & Val set \tabularnewline
    \midrule    

    \parbox[t]{2mm}{\multirow{3}{*}{\rotatebox[origin=c]{90}{no alias}}} 
    & no shift & 46.69 & 47.13 & 47.35 \tabularnewline
    & fixed shift & 47.20 & 47.38 & 47.57 \tabularnewline
    & random shift & 46.87 & 47.15 & 47.37 \\[0.5em]
    
    \parbox[t]{2mm}{\multirow{3}{*}{\rotatebox[origin=c]{90}{alias}}} 
    & no shift & 46.67 & 47.29 & 47.51 \tabularnewline
    & fixed shift & \bf{49.30} & \bf{49.25} & \bf{49.54} \tabularnewline
    & random shift & \underline{48.12} & \underline{48.44} & \underline{48.73} \tabularnewline
    \bottomrule
    \end{tabular}
    \caption{Shift and alias influence. Best PSNR in {\bf bold} and second best \underline{underlined}.}
    \label{tab:alias-shift}
\end{table}

\section{Experiments}

This section presents experiments that empirically show that Sentinel-2 imagery is well-suited for the problem of SISR.

\medskip\noindent\textbf{A surprisingly performant $L_1$ loss.}\label{subsec:real-training}
We compare two models: one trained with the $L_1$ loss and one trained with the original ESRGAN loss, with the relativistic discriminator and feature similarity terms.
Quantitatively, the average PSNR (over 12 bits) computed over the test set yields 42.21dB for the $L_1$ loss and 37.29dB for the ESRGAN loss.
Visually, we observe that the results obtained with the $L_1$ loss are slightly smoother than those obtained with the ESRGAN loss, but do not contain any color artifacts.
This can be seen in the second row of Figure~\ref{fig:results-worksite}.
In addition, the details in the images generated by the $L_1$ model are much better than those obtained through bicubic interpolation.
Overall, these experiments suggest that, in the case of Sentinel-2, the $L_1$ loss is an effective solution to increase the resolution by a factor 2 without risking the introduction hallucinated details~\cite{Blau_2018_CVPR_short}.%

Furthermore, as discussed in~\cite{zabalza2022super_short}, GAN-like losses are particularly important when LR-HR image pairs are not well registered, which is not the case for our S2/PS dataset.

Additional results using the $L_1$ model on the test set are shown in Figure~\ref{fig:results}.
We find that the network is able to resolve aliased patterns into high-frequency details, with strong fidelity to the ground-truth PlanetScope images.
Given that the $L_1$ loss minimizes distortion~\cite{Blau_2018_CVPR_short}, one can be confident that there are few hallucinated details. In ambiguous cases, the network will likely favor a blurry result instead of sharp, but potentially wrong details.

\medskip\noindent\textbf{Cross-spectral information.}\label{subsec:cross-spectral}
We claim that our network exploits cross-spectral information to increase spatial resolution. To validate this hypothesis, we perform the following experiment: from the S2/PS dataset, we train three networks, each dedicated to super-resolving one specific spectral band, and only this band is given as input.
On the test set, we observe a drop of 0.88dB in the PSNR,
and we observe visually that the network is no longer able to resolve fine structures such as very high-frequency patterns. Even though the LR signal is aliased in each spectral band, the network no longer has the ability to perform a consistent, joint reconstruction of the signal. This can be observed in the top-right image of Figure~\ref{fig:results-worksite}.
A related observation was reported in~\cite{galar2020super_short} in which a network trained with both RGB and NIR bands performed better than just with RGB bands.

\medskip\noindent\textbf{Aliasing and band-shift influence.}\label{subsec:alias-shift-influence}
We argue that the model described in Section~\ref{sec:method} is able to exploit specific characteristics of the Sentinel-2 sensor, namely the presence of alias in each band and the inter-band shifts.
The alias is due to a low spatial sampling with respect to the modulation transfer function (MTF) of the instrument~\cite{gascon2017copernicus_short}, and the inter-band shifts originate from time delays between the acquisition of the lines of the different spectral bands~\cite{gascon2017copernicus_short}.
Combined, these two aspects yield a configuration that is better-posed than standard SISR, and real information can be recovered under these acquisition specificities.
Next, we provide experimental evidence that the acquisition configuration of Sentinel-2 is indeed favorable to SISR.
To this aim, we construct synthetic datasets using six different acquisition configurations: with and without alias, and with and without fix/random inter-band shifts.
In each configuration, we use the PlanetScope images as ground-truth and we synthesize LR images according to each configuration.
The presence of alias is controlled by the amount of blur introduced before downsampling. The shifts are +/-1 offsets applied to the bands before downsampling and then compensated by 0.5 offsets on the LR images.
In each configuration, 0.1\%
The first row of Figure~\ref{fig:synthetic-dataset} shows the effects of these configurations over the generated LR images. The configuration \emph{with alias} and \emph{random inter-band shift} is the most faithful simulation of Sentinel-2 imagery.

We train one network per scenario according to the same training details as in Section~\ref{subsec:real-training}.
Table~\ref{tab:alias-shift} shows the PSNR (over 12 bits) over the train, validation and test sets for the different settings, and the bottom row of Figure~\ref{fig:synthetic-dataset} shows SR results.
These results highlight that the combined presence of the inter-band shift and the alias allows the network to retrieve significant information from the signal.
In contrast to the usual SISR scenario where little alias and no inter-band shift are present, our experiments assert that Sentinel-2 imagery is well-suited for SISR.

\section{Conclusion}

Our study aims to investigate the factors that enable SISR of Sentinel-2 imagery.
While surprising at first, our extensive experiments on carefully designed synthetic datasets show that the resolution gain can be explained by the ability of the network to exploit the characteristics of Sentinel-2 imagery, namely alias and inter-band shift.
We also validate our study by training a model for Sentinel-2 SR using a simple $L_1$ loss. Our model successfully increases the spatial resolution of Sentinel-2 images from 10m to 5m GSD, while minimizing distortion and avoiding the creation of false details.

\bibliographystyle{IEEEbib}
\bibliography{main}

\begin{thebibliography}{10}

\bibitem{Blau_2018_CVPR_short}
Y.~Blau and T.~Michaeli,
\newblock ``The perception-distortion tradeoff,''
\newblock in {\em CVPR}, 2018.

\bibitem{lanaras2018super_short}
C.~Lanaras, J.~Bioucas-Dias, S.~Galliani, E.~Baltsavias, and K.~Schindler,
\newblock ``Super-resolution of sentinel-2 images: Learning a globally
  applicable deep neural network,''
\newblock {\em ISPRS}, 2018.

\bibitem{gargiulo2019fast_short}
M.~Gargiulo, A.~Mazza, R.~Gaetano, G.~Ruello, and G.~Scarpa,
\newblock ``Fast super-resolution of 20 m sentinel-2 bands using convolutional
  neural networks,''
\newblock {\em Remote Sensing}, 2019.

\bibitem{pineda2020generative_short}
F.~Pineda, V.~Ayma, and C.~Beltran,
\newblock ``A generative adversarial network approach for super-resolution of
  sentinel-2 satellite images,''
\newblock {\em ISPRS}, 2020.

\bibitem{galar2020super_short}
M.~Galar, R.~Sesma, C.~Ayala, L.~Albizua, and C.~Aranda,
\newblock ``Super-resolution of sentinel-2 images using convolutional neural
  networks and real ground truth data,''
\newblock {\em Remote Sensing}, 2020.

\bibitem{zabalza2022super_short}
M.~Zabalza and A.~Bernardini,
\newblock ``Super-resolution of sentinel-2 images using a spectral attention
  mechanism,''
\newblock {\em Remote Sensing}, 2022.

\bibitem{michel2022sen2venmus}
Julien Michel, Juan Vinasco-Salinas, Jordi Inglada, and Olivier Hagolle,
\newblock ``Sen2ven$\mu$s, a dataset for the training of sentinel-2
  super-resolution algorithms,''
\newblock {\em Data}, vol. 7, no. 7, pp. 96, 2022.

\bibitem{salgueiro2020super_short}
L.~Salgueiro~Romero, J.~Marcello, and V.~Vilaplana,
\newblock ``Super-resolution of sentinel-2 imagery using generative adversarial
  networks,''
\newblock {\em Remote Sensing}, 2020.

\bibitem{Nguyen_2023_L1BSR}
NL. Nguyen, J.~Anger, A.~Davy, P.~Arias, and G.~Facciolo,
\newblock ``{L1BSR}: Exploiting detector overlap for self-supervised
  single-image super-resolution of sentinel-2 l1b imagery,''
\newblock in {\em CVPRW}, 2023.

\bibitem{nguyen2021self_short}
NL. Nguyen, J.~Anger, A.~Davy, P.~Arias, and G.~Facciolo,
\newblock ``Self-supervised multi-image super-resolution for push-frame
  satellite images,''
\newblock in {\em CVPRW}, 2021.

\bibitem{Wang_et_al_ESRGAN_2018_short}
Wang~X. et~al.,
\newblock ``Esrgan: Enhanced super-resolution generative adversarial
  networks,''
\newblock in {\em ECCVW}, 2019.

\bibitem{Agustsson_2017_CVPR_Workshops_short}
E.~Agustsson and R.~Timofte,
\newblock ``Ntire 2017 challenge on single image super-resolution: Dataset and
  study,''
\newblock in {\em CVPRW}, 2017.

\bibitem{jolicoeur2018relativistic_short}
A.~Jolicoeur-Martineau,
\newblock ``The relativistic discriminator: a key element missing from standard
  gan,''
\newblock {\em arXiv preprint}, 2018.

\bibitem{10.1007/978-3-319-46475-6_43_short}
J.~Johnson, A.~Alahi, and L.~Fei-Fei,
\newblock ``Perceptual losses for real-time style transfer and
  super-resolution,''
\newblock in {\em ECCV}, 2016.

\bibitem{anger2019assessing_short}
J.~Anger, C.~de~Franchis, and G.~Facciolo,
\newblock ``Assessing the sharpness of satellite images: Study of the
  planetscope constellation,''
\newblock in {\em IGARSS}, 2019.

\bibitem{gascon2017copernicus_short}
F.~Gascon et~al.,
\newblock ``Copernicus sentinel-2a calibration and products validation
  status,''
\newblock {\em Remote Sensing}, 2017.

\end{thebibliography}

\end{document}